%% file: main_ICDM.tex
\documentclass[conference]{IEEEtran}
\IEEEoverridecommandlockouts
\usepackage{cite}
\usepackage{amsmath,amssymb,amsfonts}
\usepackage{textcomp}
\usepackage{url}
\usepackage{xcolor}
\def\BibTeX{{\rm B\kern-.05em{\sc i\kern-.025em b}\kern-.08em
    T\kern-.1667em\lower.7ex\hbox{E}\kern-.125emX}}

\input{math_commands.tex}

\usepackage{mathletters}
\usepackage{stmaryrd}

\newtheorem{example}{Example}
\newtheorem{theorem}{Theorem}
\newtheorem{assumption}{Assumption}

\newtheorem{corollary}{Corollary}

\newtheorem{remark}{Remark}
\newcommand{\MASTANE}[1]{{\color{red}[{\bf Mastane: }#1]}}

\begin{document}

\title{A Risk-Averse Framework for Non-Stationary Stochastic Multi-Armed Bandits}

\author{\IEEEauthorblockN{Reda Alami}
\IEEEauthorblockA{
\textit{Technology Innovation Institute}\\
Masdar City, UAE \\
reda.alami@tii.ae}
\and
\IEEEauthorblockN{Mohammed Mahfoud}
\IEEEauthorblockA{
\textit{Montreal Institute for Learning Algorithms}\\
Montreal, Canada \\
mohammed.mahfoud@mila.quebec}
\and
\IEEEauthorblockN{Mastane Achab}
\IEEEauthorblockA{
\textit{Technology Innovation Institute}\\
Masdar City, UAE \\
mastane.achab@tii.ae}
}

\maketitle

\begin{abstract}
In a typical stochastic multi-armed bandit problem, the objective is often to maximize the expected sum of rewards over some time horizon $T$. While the choice of a strategy that accomplishes that is optimal with no additional information, it is no longer the case when provided additional environment-specific knowledge. In particular, in areas of high volatility like healthcare or finance, a naive reward maximization approach often does not accurately capture the complexity of the learning problem and results in unreliable solutions. To tackle problems of this nature, we propose a framework of adaptive risk-aware strategies that operate in non-stationary environments. Our framework incorporates various risk measures prevalent in the literature to map multiple families of multi-armed bandit algorithms into a risk-sensitive setting. In addition, we equip the resulting algorithms with the Restarted Bayesian Online Change-Point Detection (\algo{R-BOCPD}) algorithm and impose a (tunable) forced exploration strategy to detect local (per-arm) switches. We provide finite-time theoretical guarantees and an asymptotic regret bound of $\Tilde{\mathcal{O}}({\sqrt{K_T T}})$ up to time horizon $T$
with $K_T$ the total number of change-points.
In practice, our framework compares favorably to the state-of-the-art in both synthetic and real-world environments and manages to perform efficiently with respect to both risk-sensitivity and non-stationarity. 
\end{abstract}


\section{Introduction}


The field of multi-armed bandits (MAB) has emerged as a powerful framework for modeling sequential decision-making problems under uncertainty. Traditionally, MAB algorithms have primarily focused on stationary environments, where the underlying reward distributions of the arms remain unchanged over time. However, many real-world applications present a dynamic and evolving nature, where the reward distributions may undergo shifts, fluctuations, or entirely new arms may be introduced. These scenarios demand the development of advanced techniques that can effectively adapt to non-stationary environments while considering risk mitigation. The study of non-stationary MAB problems has gained significant attention in recent years, spurred by the need to address dynamic decision-making challenges in various domains such as finance, healthcare, online advertising, and recommendation systems. In these domains, an intelligent decision-making agent needs to continuously explore and exploit arms, while accounting for the inherent uncertainties associated with non-stationary environments. Moreover, merely maximizing expected rewards is often insufficient in practical settings, as it neglects the crucial aspect of risk management. While optimizing the average reward is important, it is equally crucial to consider the associated risks and uncertainties when making decisions. Risk-awareness aims to strike a balance between exploration and exploitation while accounting for the potential consequences of poor decisions.
Many variants of the classical MAB problem have been proposed for risk-aware applications.
Basically, they consist in replacing the expectation by different risk measures.
While \cite{szorenyi2015qualitative} focuses on quantiles, \cite{galichet2013exploration} and \cite{kolla2019risk}
both propose strategies relying on the estimation of the CVaR.
General bandits frameworks encompassing broad classes of risk criteria (including quantiles and CVaR) are studied in \cite{torossian2019mathcal} and \cite{cassel2018general}.
In \cite{sani2012risk}, \cite{vakili2016risk}, \cite{pmlr-v206-bhatt23b} and \cite{zimin2014generalized}, the quality
of an arm is assessed through a combination of its mean and its variance: for two arms with the same mean, the one with the smallest variance is deemed safer than the other one.
The max K-armed bandit problem is an extreme formulation of risk-aversion where the agent seeks to maximize the maximal observation collected across all time steps (see e.g. \cite{cicirello2005max, carpentier2014extreme, achab2017max}).
In \cite{maillard2013robust}, the risk-aversion is measured by means of the cumulant generating functions of the distributions of the arms;
the mean-variance approach then appears as a particular case of this method in the Gaussian scenario (i.e. when each arm is characterized by a normal distribution).

\paragraph{Notations}
The indicator function of any event $E$ is denoted by $\mathbb{I}\{E\}$
and the cardinality of any finite set $\mathcal{E}$ is denoted by $|\mathcal{E}|$.
Let $[N] = \{1,2,..,N\}$ denote the set of integers from $1$ to $N$ and $\llbracket i, j \rrbracket = \{i,i+1,..,j\}$ denote the set of integers from $i$ to $j$.
For $k\ge 1$, we denote $\Delta_k=~\{(p_1,\dots,p_k) \in \mathbb{R}_+^k\colon \sum_{i=1}^k p_i = 1 \}$ the probability simplex.
The cumulative distribution function (CDF) of a real-valued random variable $Z$ is the mapping $F(z)=\mathbb{P}(Z\le z)$ ($\forall z \in \mathbb{R}$), and we denote its generalized inverse distribution function (a.k.a. quantile function) by $F^{-1}\colon \tau \in [0, 1]\mapsto \inf\{z\in\mathbb{R}, F(z)\ge \tau\}$.

\paragraph{Key Contributions} A concise summary of our main contributions is as follows.

\begin{itemize}
    \item We introduce a multi-armed bandit framework designed to address the frequently overlooked, realistic cases involving risk-sensitive and non-stationary environments.
    \item We establish theoretical guarantees under assumptions related to false-alarm rate and detection delays: in particular, we derive near-optimal regret guarantees.
    \item We conduct reproducible experiments on both artificial and real-world datasets that accurately represent the targeted setting.
\end{itemize}



\section{A Review on Piecewise-Stationarity and Risk-Awareness in Multi-Armed Bandits}
\label{sec:background}
We present a review of the key concepts in the non-stationary risk-aware bandit problem and specify the precise (yet general) setting we operate on along with the problem-specific notation to be used throughout the paper.  
\paragraph{Multi-Armed Bandit Instance} We start by recalling the classic stochastic Multi-Armed Bandit (MAB) setting. Let us consider $A\ge 2$ probability distributions $\nu_1,\dots,\nu_A$ (with respective expected values $\mu_1,\dots,\mu_A$) associated to $A$ ``arms'' $a\in \bA := \{ 1,\dots,A \}$.
At each time $t\ge 1$, the agent has access to past history $H_{t-1}=( (A_1,R_1),\dots,(A_{t-1},R_{t-1}) )$
(by convention $H_0=(\emptyset)$), from which he derives a (possibly randomized) strategy.
Based on this strategy, the decision-maker chooses (or ``pulls'') an arm $A_t$
and then receives a random reward $X_{A_t,t}$ sampled from $\nu_{A_t}$ independently from past observations $H_{t-1}$ conditionally on $A_t$.
In the standard \emph{risk-neutral} case, the objective for the agent is to maximize the expected total amount of rewards $ \mathbb{E}\left[ \sum_{t=1}^T \mu_{A_t} \right] $ collected along $T$ (the ``time horizon'') iterations.
\paragraph{Risk-Neutral Stationary Regret} The aforementioned  problem reformulates equivalently as the minimization of the so-called \emph{expected regret} $R_T$ defined as the
difference between the performance of the \emph{optimal strategy in hindsight}, consisting in always pulling the arm with maximal mean, and the performance of the agent:
\begin{equation*}
    R_{T} = T \max_{1 \leq a \leq A} \mu_a - \mathbb{E}\left[ \sum_{t=1}^T \mu_{A_t} \right] \, .
\end{equation*}
\subsection{Piece-wise Stationary Bandits}
\paragraph{Switching-Bandit Problem}First defined in \cite{garivier2011upper}, the \textit{switching-bandit} problem defines a setting where \textit{abrupt changes} may happen to the environment at unknown times that we refer to as \textit{change-points}.
We call ``change-point'' a time step at which at least the distribution of one arm has changed.
We therefore concisely define a switching-bandit instance as the sequence of couples $\mathbf{B} = \left( c_k, \left( \nu^k_a \right)_{a=1}^{A}\right)_{k=1}^{K_T} $,
where $K_T$ is the
total number of changes.
For $k\in [K_T]$, $1<c_k<T $ represents the time at which occurs the $k$-th change point
(with the conventions $c_0 = 1$ and $c_{K_T+1} = T $),
and $\nu^k_a$ is the distribution of the arm $a$ at rounds
$c_{k-1} \le t < c_{k}$.
Here, we highlight that the decomposition with respect to all switches is performed in a somewhat macroscopic fashion. For instance, if change-point $c_k$ occurs locally on arm $a$ only, we would still decompose all other arms around point $c_k$, i.e into $\nu_{a^\prime}^k$ and $\nu_{a^\prime}^{k+1}$ for all $a^\prime \neq a$.

Given our approach to deal with non-stationarity, we introduce the assumption that the arms' distributions originate from a Bernoulli prior (\cite{RBOCPDTS, kaufmann2012bayesian}).
While this assumption seem quite restrictive at first glance, we cite multiple applications in a variety of fields where problems of this nature arise. In particular, monitoring the performances of statistical models, modelling the collisions in cognitive radio, monitoring events in probes for network supervision, clinical trials and recommender systems, to name a few.

\begin{assumption}[Bernoulli rewards]
The rewards of all the arms are sampled from Bernoulli distributions denoted as $ \nu_{a,t} = \bern{\mu_{a,t}} , \forall a \in \bA, \forall t \in [T]$.
For $c_{k-1} \le t < c_{k}$, these distributions are stationary:
$\mu_{a,t} = \mu^k_{a}$ and
$\bern{\mu_{a,t}}=\nu^k_a$.
\label{assump::Bernoulli}
\end{assumption}

\begin{assumption}[Change-point detectability]
\label{assump::CPDetec}
There exists a constant $\lambda > 0$ such that $\forall a \in \bA$ and $\forall t \in [T]$, if $\mu_{a,t} \neq \mu_{a, t+1}$ then $\abs{\mu_{a, t}-\mu_{a, t+1}} \geq \lambda$.
\end{assumption}

\paragraph{Per-Arm Switches}While the global switch case is an important one to consider when dealing with non-stationarity, it is quite unrealistic in practice. In our setting, we consider the more realistic local or per-arm switch case. This gives the problem a more realistic flavor and requires more subtle approaches to balance the \emph{exploration-exploitation} tradeoff.

\subsection{Risk-Aware Stationary Bandits}


In various \emph{risk-aware} contexts, the quantities of interest are not necessarily expectations.
In environmental or financial applications for instance, a decision maker may be \emph{risk-averse} by taking decisions to ensure sufficient protection against disastrous events such as flooding or financial crisis (see \cite{beirlant2006statistics}, \cite{resnick2007heavy}).
In other words, efficient strategies are designed by giving more importance to worst-case scenarios than to ``normal'' observations.
Mathematically, such pessimistic scenarios are often modeled as rare and extreme events.
Hence, in many risk-aware problems, the learner has to optimize a criterion based on the tail of some distribution, which characterizes its extreme values.
%
%
\paragraph{Risk Measures}
Several risk measures have been proposed in the literature to replace the expectation; we recall next two examples of popular risk-measures that have been considered in the risk-aware bandits literature.

\begin{example}
    \label{ex:mv}
    The mean-variance (MV) criterion is defined as follows:
    \begin{equation*}
        MV(\nu) = \mu - \gamma \sigma^2 \, ,
    \end{equation*}
    where $\mu,\sigma^2$ respectively denote the mean and the variance of some distribution $\nu$ and $\gamma \ge 0$.
    For $\gamma >0$, if two distributions $\nu_1,\nu_2$ have same mean $\mu_1=\mu_2$, the MV objective favors the one having the smallest variance.
\end{example}

\begin{example}
    \label{ex:cvar}
    The conditional value-at-risk (``CVaR'' in short, also referred as ``expected shortfall'' or ``superquantile'' \cite{rockafellar2000optimization}) at some level $0 < \alpha \le 1$ of a distribution $\nu$ with CDF $F$ is given by:
    \begin{equation*}
        \text{CVaR}_\alpha(\nu) = \frac{1}{\alpha} \int_{\tau=1-\alpha}^1 F^{-1}(\tau) d\tau \ .
    \end{equation*}
    Intuitively, the CVaR is a best-case average of the distribution $\nu$ over the ratio $\alpha$ of largest quantiles.
\end{example}

Note that in the limit cases $\gamma=0$ and $\alpha=1$, the two risk measures described in Examples \ref{ex:mv}-\ref{ex:cvar} coincide with the mean.

The CVaR$_\alpha$ (with $\alpha << 1$) of some distribution $\nu$ allows to study its right tail; empirically, by selecting the $\alpha$-fraction of the largest order statistics $X_{\sigma(\lfloor (1-\alpha) T \rfloor)},\dots,X_{\sigma(T)}$ among an i.i.d. sample $(X_t)_{1\le t\le T}$ (see Eq. (\ref{intro_e_eq:estim_cvar})).
Empirically, the CVaR$_\alpha$ of some distribution $\nu$ can be estimated from an i.i.d. sample $X_t\sim \nu$ for $1\le t\le T$ by:
\begin{equation}
\label{intro_e_eq:estim_cvar}
\widehat{\text{CVaR}}_\alpha = \frac{1}{\lceil \alpha T \rceil} \sum_{t=\lfloor (1-\alpha)T \rfloor}^T X_{\sigma(t)},
\end{equation}
with permutation $\sigma\in\mathfrak{S}_T$ and the order statistics $X_{\sigma(1)}\le \dots\le X_{\sigma(T)}$.
We point out that $\widehat{\text{CVaR}}_\alpha$ is an $L$-statistic (see \cite{van2000asymptotic}), i.e. a linear combination of the order statistics.

\paragraph{The Risk-LCB Algorithm}


In this paper, we study the Multi-Armed Risk-Aware Bandit (MARAB) in the case of a non-stationary environment.
In stationary MARAB, the quantity of interest in not the expected value anymore, but rather a risk
measure $\rho$. Hence, the goal in MARAB is to minimize the expected $\rho$-regret given by
\begin{equation}
    \label{eq:rho_regret}
    R_{T}^{\rho} := \mathbb{E}\left[\sum\limits_{t=1}^T\rho(\nu_{A_t})\right]-T\min\limits_{1 \leq a \leq A}\mathbb{E}\left[ \rho(\nu_a)\right]
\end{equation}
with $\rho^\star = \min_a \rho(\nu_a) $ being the risk-measure of the optimal arm $a^* = \argmin_a \rho(\nu_a)$
assumed to be unique for simplicity.

We denote $\rho_n=\rho(\nu_n)$ the risk measure of an empirical measure $\frac{1}{n} \sum_{s=1}^n X_s$
made of $n\ge 1$ i.i.d. samples.
Following \cite{tan2022survey}, we assume that the risk measure of interest satisfies a Lipschitz continuity property.
\begin{assumption}
    \label{ass:lipschitz}
    There exists a Lipschitz constant $L$ such that for any pair of distributions $\nu_1,\nu_2$ over the real line,
    \begin{equation*}
        | \rho(\nu_1) - \rho(\nu_2) | \le L \int_{\tau=0}^1 | F_1^{-1}(\tau) - F_2^{-1}(\tau) | d\tau \, .
    \end{equation*}
\end{assumption}
For instance, the conditional value-at-risk described in Example \ref{ex:cvar} satisfies Assumption \ref{ass:lipschitz} for the Lipschitz constant $L=\frac{1}{\alpha}$ (the proof easily follows from a triangular inequality).

\begin{remark}
It is important to highlight that \textit{Assumption \ref{ass:lipschitz}} is of utmost importance to the non-stationary setting. To see that, imagine that a given switching-bandit problem satisfies \textit{Assumption \ref{assump::CPDetec}}. If a non-Lipschitz risk measure $\rho$ is used, there is no way to ensure that even a small change in the arms' means does not correspond to a drastic one in the risk-aware setting measure by $\rho$.
\end{remark}

We recall that the \algo{Risk-LCB} strategy introduced in \cite{tan2022survey} pulls the arm $A_t$ minimizing some lower confidence bound index. More precisely, the algorithm chooses the arm $A_t = \argmin_{1\le a\le A} \Index{a}{\algo{Risk-LCB}}{t}$ with \algo{Risk-LCB} index given by
\begin{equation}
    \label{eq:index_lcb}
    \Index{a}{\algo{Risk-LCB}}{t} = \rho_{a, N_a(t-1)} - w_{a, N_a(t-1)}
\end{equation}
where $\omega_{a, N_a(t-1)} = \frac{L \sigma [ 32 \sqrt{e \log (t)} + 512 ] }{\sqrt{N_a(t-1)}}$ represents the upper confidence bonus favoring exploration and $N_a(t-1) =\sum_{s=1}^{t-1}  \mathbb{I}_{\left\{A_s=a\right\}}$ is the number of times arm $a$ has been played up to time $t-1$.

\paragraph{Risk Aware Stationary Regret Guarantees} By leveraging the concentration properties of $L$-Lipschitz risk measures for sub-Gaussian random variables, the following risk-aware regret bound has been derived in \cite{tan2022survey}.

\begin{theorem}{\textsc{($\rho$-Regret bound, \cite{tan2022survey})}.}
    \label{thm:regret_riskLCB}
    For an $L$-Lipschitz risk measure $\rho$ and $A$ arms with sub-Gaussian distributions $\nu_a$ ($1\le a\le A$) of parameter $\sigma^2$, the expected $\rho$-regret of Risk-LCB on $T$ rounds is upper bounded as follows
    \begin{equation*}
        R^\rho_T \le \sum_{a\neq a^*} \frac{4L^2\sigma^2[32\sqrt{e \log(T)}+512]^2}{\Delta_a} + 28 A \: \Delta_a \ , 
    \end{equation*}
where $\Delta_a = \rho(\nu_a) - \rho(\nu_{a^\star})$.

\end{theorem}

\subsection{Incorporating Non-Stationarity \& Risk-Awareness}
By combining the previous sections, we can now introduce the risk-aware $\rho$-regret for a non-stationary switching-bandit problem $\mathbf{B} = \left(c_k, \left(\nu^k_a \right)_{a=1}^{A}\right)_{k=1}^{K_T} $:


\begin{equation}    \mathcal{R}_{T}^\rho(\mathbf{B}) = \mathbb{E}\left[\sum\limits_{t=1}^T\rho(\nu_{A_t, t})\right]
- \sum\limits_{k=1}^{K_T+1} (c_{k}-c_{k-1}) \min_{1\leq a \leq A} \mathbb{E}\left[\rho(\nu^{k}_a)\right] 
\end{equation}

\section{A Change-Point Detection Approach to Modeling Non-Stationarity}

Here, we present the motivation behind our choice of adopting a change-point detection approach along with its details and performance guarantees.



The authors in \cite{RBOCPD} have designed a variant of the original Bayesian online change-point detector introduced by  \cite{adams2007bayesian}. The resulting strategy is named restarted Bayesian online change-point detector \algo{R-BOCPD}. It is a pruning version of the original algorithm reinterpreted from the standpoint of forecasters aggregation and expressed as a restart procedure pruning the useless forecasters. 

More formally, for a binary sequence $\tuple{ z_r,..., z_n } \in \{0,1\}^{n-r+1}$, the final formulation of the \algo{R-BOCPD} strategy takes the following form
\begin{align}
\algo{R-BOCPD}(z_r,...,z_t) =  \indFct{ \exists s \in \segOL{r,t}:\vartheta_{r,s,t} >  \vartheta_{r,r,t}}
\end{align}

where the weight of the forecasters $\vartheta_{r,s,t}$ are computed in a recursive way as follows (assuming an initial weight $\vartheta_{r,1,1}=1$, and a temporal function $\eta_{r,s,t} \in (0,1)$) such as
\begin{align}
\label{eq::weight}
   \vartheta_{r,s,t} &= \begin{cases}
		\frac{\eta_{r,s,t}}{\eta_{r,s,t-1}}	 \exp \left(-l_{s,t} \right) \vartheta_{r,s,t-1} & \forall s<t,\\
			\eta_{r,t,t} \times \cV_{r:t} & \text{if }s=t\, ,
		\end{cases}
\end{align}
where the initial weight of the forecaster takes the form $\cV_{r:t} \define  \exp \left(-\sum_{s'=r}^{t-1} l_{s',t-1}\right)$ and the instantaneous loss $l_{s,t} \define  -\log \lap{z_{t}}{z_{s}...z_{t-1}}$ is computed based on the Laplace predictor: 
$
    \lap{z_{t}}{z_{s}...z_{t-1}}  \define \begin{cases}
	\frac{\sum_{i=s}^{t-1} z_i + 1 }{t-s + 2 } &  \text{if } z_{t} = 1 \\
	\frac{\sum_{i=s}^{t-1}(1- z_i) + 1 }{t-s+ 2 }  & \text{if }  z_{t} = 0
	\end{cases}
$. 
The hyper-parameter $\eta_{r,s,t}$ is tuned as a decreasing function in $t$ and depends also on the probability of false alarm $\delta$.

\paragraph{Choice of R-BOCPD}\algo{R-BOCPD} is chosen among all the sequential change-point detector algorithms in the state of the art for three main reasons.

\begin{itemize}
    \item \textit{Well adaptability to unknown priors.} Indeed, the \algo{R-BOCPD} algorithm has been  designed to solve the problem of sequential change-point detection in a setting where both the
change-points and the distributions before and after the change are assumed to be unknown. This setting corresponds exactly to the situation of an agent facing a multi armed bandit whose distributions are unknown and may change abruptly at some unknown instants. 
    \item \textit{Minimum detection delay.} This corresponds to the first criteria assessing the performance of a sequential change-point detector. The detection delay is defined as the number of samples needed to detect a change. In \cite{RBOCPD}, the authors have shown that the detection delay of the \algo{R-BOCPD} strategy is asymptotically optimal in the sense that it reaches the existing lower bound stated in Theorem 3.1 in \cite{lai2010sequential}.
    
    \item \textit{Well controlled false alarm rate.} The false alarm rate corresponds to the probability of detecting a change at some instant where there is no change. Again, in \cite{RBOCPD}, the authors have demonstrated that $\forall \delta \in (0,1)$ \algo{R-BOCPD} doesn't make any false alarm with a probability at least $1-\delta$. 
\end{itemize}

\section{A Non-Stationary Risk-Aware Bandit Setting}


Now, we choose to equip our risk aware bandit strategy (here \algo{Risk-LCB}) with the \algo{R-BOCPD} approach. We present the details of the introduced algorithm along with its theoretical guarantees and the choice of parameters controlling its performance.

\subsection{\algo{R-BOCPD}-equipped \algo{Risk-LCB}}
Our strategy consists in equipping each arm $a$ in the $A$-armed switching-bandit problem. To model per-arm switches, we add a \textit{forced exploration} parameter to sample undersampled arms. Our algorithm is as follows
\begin{algorithm}[H]
	\caption{\algo{R-BOCPD-Risk-LCB} for risk-averse $A$-armed switching bandits with Bernoulli means} 
	\label{alg::1}
	\begin{algorithmic}[1]
	\Require{$\bA$: Arm set, $\textsc{Bandit}$: Multi-Armed Risk-Aware Bandit strategy as subroutine, $\beta \in (0,1)$: forced exploration rate, $s_0 > 0, n_0 > 0$: parameters for initialization, $T$: Horizon.}

\Initialize{$\forall a \in \bA \quad  \N{a}{0} = n_0 \text{ and } \Su{a}{0} = s_0$}
		\Define{
		\vspace{-2em}
		\begin{align}
		    &\forall a \in \bA, \forall t \in \bT: \quad\Index{a}{Bandit}{t} \text{ is defined following some risk}\nonumber \\ &\text{ averse MAB strategy (e.g. \algo{Risk-LCB}).} \label{eq::Bandit}
		\end{align}
		}
		\For{$t=1,\dots, T$}
		\State Choose action $A_t =
		\begin{cases}
		\argmin_a \Index{a}{Bandit}{t} & \text{with probability } 1-\beta \\
		a & \forall a \in \bA \ \text{with probability } \frac{\beta}{A}
		\end{cases}
		$.
		\State Observe $X_{A_t,t} \sim \bern{\mu_{A_t,t}}$.
		\State Re-shift observation $Y_{A_{t}, \N{A_t}{t} } = X_{A_t,t}$.
		\State Update $\N{A_t}{t+1} = \N{A_t}{t}+1$ and $\Su{A_t}{t+1} = \Su{A_t}{t} + X_{A_t,t}$.
		\State Perform change-point detection using \algo{R-BOCPD} on the sequence $\left(Y_{A_t,1},..., Y_{A_t, \N{A_t}{t}} \right)$. 
		\If{$\CPD{Y_{A_t,1},..., Y_{A_t, \N{A_t}{t}}} = 1$}
		\State $\N{A_t}{t+1} = n_0 \text{ and } \Su{A_t}{t+1} = s_0$
		\EndIf
		\State Update $\Index{A_t}{Bandit}{t+1}$ according to Eq.(\ref{eq::Bandit}).
		\EndFor
	\end{algorithmic}
\end{algorithm}

\subsection{Near-Optimal Regret Bound} 
Given the theoretical guarantees that come naturally with \algo{R-BOCPD}, i.e the minimal detection delay and false-alarm rates, we can formulate a bound on the number of suboptimal choices of a given action $a$ in the following Theorem \ref{thm:Na}. This is directly used to estimate the theoretical guarantees of \algo{R-BOCPD-Risk-LCB} in the Corollary \ref{corr:regret}.

\begin{theorem}[Bounding the number of samples related to sub-optimal arms]
\label{thm:Na}

Consider the random variable $\overline{N}_{a,T} = \sum_{t=1}^T \mathbb{I}\{ A_t = a \neq a_t^\star \}$ the number of times arm $a$ is pulled until time $T$ while not being the optimal arm, then under Assumptions \ref{assump::Bernoulli} and \ref{assump::CPDetec}, for any $\beta \in (0,1)$ and any arm $a \in \bA$, the \algo{R-BOCPD-Risk-LCB} strategy achieves:
\begin{align*}
 \esp{\overline{N}_{a,T}} \leq \frac{\alpha T}{A} & + \sum_{k=1}^{K_T} \esp{D_{a,k}} +  \left(K_T + \esp{F_T} \right)\times 
\\
&\left[ \frac{4L^2\sigma^2[32\sqrt{e \log(T)}+512]^2}{ \min\limits_{k \in [1,K_T], a\neq a_{[k]}^\star} \Delta_{a, [k]}^2} + 28 A \right] \ .
\end{align*}
where $D_{a,k}$ denotes the detection delay of arm $a$ and $F_T$ denotes the false alarm rate.
\end{theorem}

\begin{corollary}[Finite-time Regret Bound]
\label{corr:regret}
Under Assumption \ref{assump::Bernoulli}, by choosing a fixed $\beta = \sqrt{\frac{A K_T}{T}}$, the regret upper bound of the strategy Bayesian Change-point detection using Risk-LCB takes the following form:
\vspace{-1em}
\begin{align*}
    \mathcal{R}^{\algo{RBOCPD-RiskLCB}}_T = \mathcal{O}\left( \frac{K_T L^2\sigma^2e \log(T)}{ \min\limits_{k \in [1,K_T], a\neq a_{[k]}^\star} \Delta_{a, [k]}^2} + \sqrt{A K_T T}  \right)
\end{align*}
\end{corollary}

\subsection{Forced Exploration \& Choice of $\beta$}
We recall from {\ref{assump::CPDetec} that one key pre-requisite for dealing with non-stationarity is that the change-points that happen to switching-bandit are detectable. Within the \algo{R-BOCPD} framework, it is equivalent to ensuring that the \textit{forced exploration} strategy ensures that each arm $a$ is provided enough samples for its equipped \algo{R-BOCPD} to detect its change-points. Given the Lipschitz assumption on the risk measures, the change-points we miss given a small change in the means (not satisfying \textit{Assumption} \ref{assump::CPDetec}) would not be dratically more risky and thus not contribute much to the regret. 

\subsubsection{Global Switch Case}
A global switch case defines the case where a change-point  is synchronized across arms. It arises in many problems practical where the arms' behavior is controlled by a centralized entity. When $\alpha=0$, our algorithm simplifies to efficiently address this problem.
\subsubsection{Handling Local Switch Scenarios}
This case is much more complicated to model than the former rather simple global switch case. Our introduced algorithm controls the \textit{exploration-exploitation} tradeoff directly via the parameter $\alpha$. The order-optimal choice of a \textit{fixed} $\beta$ to minimize the regret expression in Theorem \ref{thm:Na} is $\beta = \sqrt{\frac{A K_T}{T}}$. This requires the à-priori knowledge of both $T$ and $K_T$, which is not the case in practice. One example of a strategy for tuning $\beta$ is the following: given the Bernoulli assumption on the arms' reward means, we can approximate the change-point arrival process by a Poisson process whose intensity can be set to some decreasing function of time, for instance. We highlight that this process only controls the exploration strategy so that the \algo{R-BOCPD} at each arm is given enough samples to be able to decide if there was a change-point or not. We propose a more elaborate analysis of the exploration strategy in the Appendix, due to the space constraints.


\subsection{Proof Sketch}

Let us consider the random variable $\overline{N}_{a,T} = \sum_{t=1}^T \mathbb{I}\{ A_t = a \neq a_t^\star \}$ the number of times arm $a$ is pulled until time $T$ while not being the optimal arm. Our regret analysis mainly consists in controlling the expected value of $\overline{N}_{a,T}$. The proof derives from the following steps.
\paragraph{Step 1: Decomposition of $\mathbb{E}[\overline{N}_{a,T}]$}
Given that we are dealing with a piecewise stationary multi-armed bandit problem, we can decompose the total regret across the different stationary periods and the time of change-points detection. In particular, we can write $\mathbb{E}[\bar{N}_{a,T}]$ as a function of:
\begin{itemize}
    \item the expected delays $\mathbb{E}[D_{a,k}]$ of change-point detection,
    \item the expected number $\mathbb{E}[F_T]$ of false alarms,
    \item the risk-averse regret on each stationary period.
\end{itemize}

\paragraph{Step 2: Controlling detection delays and false alarms}
We build upper bounds over the expected detection delays $\mathbb{E}[D_{a,k}]$ and the expected number of false alarms $\mathbb{E}[F_T]$ of \algo{R-BOCPD}.

\paragraph{Step 3: Stationary regrets}
We can upper bound the regret on each stationary period using the regret bound of the risk-aware bandit algorithm Risk-LCB recalled in Theorem \ref{thm:regret_riskLCB}. This bound takes into account the risk aversion of the decision maker through the risk-measure $\rho$ and depends on the corresponding suboptimality gaps.


\subsection{Detailed proofs}


In this section, we construct the theoretical guarantees of our algorithmic framework in continuation of Section 4.4 but in more detail, invoking the near-optimal theoretical advantages of \algo{R-BOCPD} in our setting and developing regret guarantees of risk-averse nature. Given the natural correspondence between the regret of a $A$-armed bandit problem and the number of choices of some action $a \in \bf{A}$ where $a$ is sub-optimal, we first attempt to bound this quantity. More formally we define 
\begin{align*}
    \esp{\overline{N}_{a,T}} = \sum\limits_{t = 1}^T \indFct{a_t = a \neq a^\star_t}
\end{align*}
where we recall that $a^\star_t$ is the optimal arm at time $t$ and $a_t$ is the minimizer of some (risk-aware) lower confidence bound index, i.e $a_t = \argmin_{1\le a\le K} \Index{a}{\algo{Risk-LCB}}{t}$. 
\subsubsection{Bounding $\esp{\overline{N}_{a, T}}$}
Since we consider the general per-arm switch setting, our algorithm needs to incorporate a \textit{forced exploration} component to be able to detect changes happening in only a subset of arms, which might be undersampled by the (exploitation) policy. For a $\bf{A}$-arm bandit instance, we choose an arm uniformly at random $\beta T$ times, where $\beta$ is a tunable parameter, which may also depend on $t$ in general. Now, we are ready to construct the regret guarantees of \texttt{R-BOCPD}-equipped \texttt{Risk-LCB}. First, we start by writing the probability that a sub-optimal arm $a$ is played at time instance $t$

\begin{align*}
    \pr{A_t = a \neq a^\star_t} \leq \frac{\beta}{A} + (1-\beta)\:\pr{a_t = a \neq a^\star_t}
\end{align*}

Recalling that $\overline{N}_{a, T}$ is defined as the number of scenarios where sub-optimal action $a$ was chosen up to time horizon $T$, we bound the expected value of the latter quantity as follows

\begin{align*}
    \esp{\overline{N}_{a,T}} & \leq \:\sum_{t=1}^T \pr{A_t = a \neq a^\star_t} \\
    & \leq \:\sum_{t=1}^T \left( \frac{\beta}{A} + (1-\beta)\:\pr{a_t = a \neq a^\star_t}\right) \\
    & \leq  \frac{\beta}{A}T +\underbrace{ \sum_{t=1}^T  \pr{a_t = a \neq a^\star_t}}_{(1)}
\end{align*}

Now, we need to construct an upper bound for the term in (1). We first recall the setting \algo{Risk-LCB} operates on. At initialization, it first samples all arms once in the first $A$ rounds, then it selects the action with the lowest risk index $\Index{a}{\algo{Risk-LCB}}{t}$ for $t \geq A+1$ with probability $\frac{1-\beta}{A}$ while balancing that with an exploration strategy uniform over all arms with probability $\beta/A$. Introducing the local (per-arm) switch assumption, we equip each arm $a \in \bf{A}$ with \algo{R-BOCPD}. To derive the theoretical regret guarantees, we need to first decompose the time horizon $T$ with respect to the change-points in each arm $a$. For that, we first define a function $\tau_a(t)$ that computes the last \algo{R-BOCPD} restart for arm $a$ prior to time $t$. Note that this restart may also be a false-alarm. Finally, we define the set of piecewise stationary bandit instances as defined by \algo{R-BOCPD}'s detected change-points (including false-alarms) for some action $a \in \bf{A}$ as follows

\begin{align*}
     \mathcal{T}_a = \left\lbrace t \in \llbracket1,T\rrbracket : \forall s \in \llbracket\tau_a(t)+1, t\rrbracket, \mu_{a,s} = \mu_{a,t} \: \right\rbrace   
\end{align*}
Now, we adopt a decomposition with respect to \algo{R-BOCPD}'s detected change-points, which again may include change-points. During the detection phase, we assume our algorithm assumes a maximum instantaneous regret of $1$. Defining the set of change-points for action $a$ as $\left\{c_{\ell}^{a}\right\}_{\ell=1}^{K_{T}^{a}}$ where $c_0 = 0$ and $c_{K_{T}^{a}+1} = T + 1$. We consequently write the bound for term (1) as 

\begin{align*}
    \sum_{t=1}^T & \pr{a_t = a \neq a^\star_t} 
     \leq \esp{\sum_{\ell = 1}^{K_{T}^{a}} D_{a, \ell} + \sum_{t \in \mathcal{T}_a} \indFct{a_t = a \neq a^\star_t}} \\
    & \leq \sum_{\ell = 0}^{K_{T}^{a}} \left\{\esp{D_{a, \ell}} + R_{[c_{\ell+1}^{a}-c_{\ell}^{a}-\esp{D_{a, \ell}}, c_{\ell+1}^{a}]}^{\rho}\right\}
    \\
    & \overset{(2)}{\leq} \sum_{\ell=0}^{K_{T}^{a}}  \frac{4L^2\sigma^2[32\sqrt{e \log( c_{\ell+1}^{a}-c_{\ell}^{a}-\esp{D_{a, \ell}})}+512]^2}{\Delta_{a, [\ell]}^2}
    \\
    & + \sum_{\ell=0}^{K_{T}^{a}} \left\{\esp{D_{a, \ell}} + 28K\right\}
\end{align*}

where $D_{a, \ell}$ corresponds to the detection delay of change-point $\ell$ ($D_{a, 0} = 0$) occuring to arm $a$ and (2) results from the regret bound of \algo{Risk-LCB}. We emphasise that, given the non-stationary setting, the regret definition is indexed by interval to be more precise (while the bound obviously remains similar for intervals of similar lengths). Since the regret is only logarithmic in the length of the piecewise-stationary periods, we can trivially bound it by the overall time horizon $T$. Finally, we write 
\begin{align*}
        \sum_{t=1}^T & \pr{a_t = a \neq a^\star_t} 
     \leq \sum_{k=0}^{K_{T}^{a}} \esp{D_{a,k}} + \left(K_{T}^{a} + \esp{F_{T}^{a}} \right)\times \\
     &
\left[ \frac{4L^2\sigma^2[32\sqrt{e \log(T)}+512]^2}{ \min\limits_{k \in [1,K_{T}^{a}], a\neq a_{[k]}^\star} \Delta_{a, [k]}^2} + 28 A \right] \
\end{align*}
where $\Delta_{a, [k]} = \rho(\nu_{a}^k) - \rho(\nu_{a^\star}^k)$ and $\esp{F_{T}^{a}}$ corresponds to \algo{R-BOCPD}'s expected false-alarm rate for arm $a$ up to time horizon $T$, which can be bounded by $\delta$ with high probability. Consequently
\begin{align}\label{subop}
     \esp{\overline{N}_{a,T}} 
     & \leq  \frac{\beta}{A}T + \sum_{k=0}^{K_{T}^{a}} \esp{D_{a,k}} + \left(K_{T}^{a} + \esp{F_{T}^{a}} \right) \nonumber \\ &\times 
\left[ \frac{4L^2\sigma^2[32\sqrt{e \log(T)}+512]^2}{ \min\limits_{k \in [1,K_{T}^{a}], a\neq a_{[k]}^\star} \Delta_{a, [k]}^2} + 28 A \right] \\
     & \leq \frac{\beta T}{A} + \sum_{k=1}^{K_T} \esp{D_{a,k}} + \left(K_T + \esp{F_T} \right) \nonumber \\ & \times 
\left[ \frac{4L^2\sigma^2[32\sqrt{e \log(T)}+512]^2}{ \min\limits_{k \in [1,K_T], a\neq a_{[k]}^\star} \Delta_{a, [k]}^2} + 28 A \right] \ 
\end{align}

where $K_T = \sum\limits_{a \in \bf{A}} K_{T}^{a}$ and $\esp{F_T} = \max\limits_{a \in \bf{A}} \esp{F_{T}^{a}}$

\subsubsection{Proof of \algo{R-BOCPD}'s Theoretical Guarantees- Optimal False-alarm Rate and Detection Delay}
\label{Sec::BOCPD_proof}
Regarding the false alarm control, it comes directly from Theorem 1 in the analysis of the \algo{R-BOCPD} in \cite{RBOCPD}.

Indeed, we have:

\begin{align*}
\forall \delta' \in (0,1) : 	
&	\esp{F_T} \leq \sum_{k=1}^{K_T}
 \mathbb{P} \Big(
	  t \in \segOR{\tau_k+1, \tau_{k+1}-1}: \\ &  \CPD{Y_{A_t,1},..., Y_{A_t, \N{A_t}{t}}} = 1 \Big) \\
	&\leq K_T \delta'.
	\end{align*}

Thus, by choosing $\delta' = \frac{\delta}{K_T}$, we upper bound $\esp{F_T} \leq \delta$.

Then, the control of the detection delay comes also from theorem 2 in the analysis of the restarted Bayesian online change-point detector in \cite{RBOCPD}.

Indeed we upper bound the detection delay of change point $\tau_{a,k}$ related to arm $a$ (with some $\delta' \in (0,1)$)

\begin{align}
\esp{D_{a,k}} =  \min \Bigg\{ d \in \mathbb{N}^\star: & d > \frac{\left(1 - \frac{ \cC_{\tau_{a,k}, d+\tau_{a,k}-1,\delta}}{\Lambda_{a,\left[k\right]}}  \right)^{-2}}{2 \Lambda_{a,\left[k\right]}^2 }\times \nonumber \\ 
&  \frac{-\log \eta_{\tau_{a,k},d+ \tau_{a,k}-1} + f_{\tau_{a,k},d+ \tau_{a,k}-1}}{1+\frac{\log \eta_{\tau_{a,k},d+ \tau_{a,k}-1} - f_{1,\tau_{a,k},d+ \tau_{a,k}-1}}{2  n_{1:\tau_{a,k}-1} \left(\Lambda_{a,\left[k\right]} - \cC_{\tau_{a,k}, d+\tau_{a,k}-1,\delta}  \right)^2}} \Bigg\}, \label{eq::detectDelay}
\end{align}
where
\begin{align}
\cC_{s,t,\delta} = & \frac{\sqrt{2}}{2} \Bigg( \sqrt{\frac{1+\frac{1}{n_{1:s-1}}}{n_{1:s-1}} \log \left(\frac{2\sqrt{n_{1:s}}}{\delta'}\right)} + \nonumber
 \\ 
 & \sqrt{\frac{1+\frac{1}{n_{s:t}}}{n_{s:t}} \log \left(\frac{2n_{1:t} \sqrt{n_{s:t}+1} \log^{2}\left(n_{1:t}\right)}{\log(2) \delta'}\right)}\Bigg). \label{Eq::C}
\end{align}
 
with $f_{s,t} = \log n_{1:s}  + \log  n_{s:t+1}  - \frac{1}{2} \log n_{1:t} + \frac{9}{8}$ and  the decreasing function $n_{i:j} = j-i+1$ and $\eta \in (0,1)$.

Indeed assuming that we collect enough samples between two consecutive change-points, we upper bound the detection delay of change point $\tau_{a,k}$ related to arm $a$ by its behavior in the asymptotic regime such that:

\begin{align*}
    \esp{D_{a,k}} = \mathcal{O}\left( \frac{\smallO{\log \tfrac{1}{\delta'}}}{2 \beta \times \Lambda_{a,\left[k\right]}^2} \right) \leq \mathcal{O}\left( \frac{\smallO{\log \tfrac{1}{\delta'}}}{2 \beta \times \min \limits_{a : \Lambda_{a,\left[k\right]} \neq 0)} \Lambda_{a,\left[k\right]}^2} \right)  
\end{align*}

Finally, by choosing $\delta' = \frac{\delta}{K_T}$ we get the following result:

\begin{align}
    \esp{D_{a,k}} \leq \mathcal{O}\left( \frac{\smallO{\log \tfrac{K_T}{\delta}}}{2 \beta \times \min \limits_{a : \Lambda_{a,\left[k\right]} \neq 0)} \Lambda_{a,\left[k\right]}^2} \right)  
    \label{Eq::Delay}
\end{align}

\subsubsection{Finite-time Regret Bound}
By virtue of \algo{R-BOCPD}'s optimal expected false-alarm and detection delay guarantees (see subsection \ref{Sec::BOCPD_proof}) and fixing an exploration strategy parametrized by a fixed $\beta = \sqrt{\frac{A K_T}{T}}$, then by injecting the result of Equation \ref{Eq::Delay} into Equation \ref{subop} and after summing over all the arms, we write the order-optimal regret of \algo{R-BOCPD}-equipped \algo{Risk-LCB} bound as: 
\begin{align*}
    \mathcal{R}^{\algo{R-BOCPD-Risk-LCB}}_T = \mathcal{O}\left( \frac{4 K_T L^2\sigma^2e \log(T)}{ \min\limits_{k \in [1,K_T], a\neq a_{[k]}^\star} \Delta_{a, [k]}^2} + \sqrt{A K_T T}  \right)
\end{align*}
which proves Corollary 1. 



\section{Numerical Experiments}
 We evaluate our algorithm in practice in both synthetic and real-world environments. We list the details of the setting, the experimental setup and results in the following.

\subsection{Setting and baselines}
We present the details of \algo{R-BOCPD-Risk-LCB} in practice and choose to benchmark it against 2 algorithms that perform the best within the non-stationary bandit setting that are considered as baselines for the non-stationary risk aware setting.
\begin{itemize}
    \item \algo{R-BOCPD-Risk-LCB}: \algo{Risk-LCB} equipped with \algo{R-BOCPD} as explained in the framework of Algorithm \ref{alg::1} where the index of arm $a$ at time $t$ $\Index{a}{Bandit}{t}$ is replaced by the index of Risk LCB $\Index{a}{\algo{Risk-LCB}}{t}$ defined in Equation (\ref{eq:index_lcb}).
    \item \algo{Discounted Risk LCB}: Inspired from the Discounted UCB algorithm in \cite{garivier2011upper}, the index of Discounted Risk LCB for action $a$ at time $t$ takes the following form:
    $
        \Index{a}{\algo{Discounted Risk-LCB}}{t} = \rho_{a, N^\gamma_a(t-1)} - w_{a, N^\gamma_a(t-1)}
    $
    where $\rho_{a, N^\gamma_a(t-1)}$ is the risk measure of the discounted empirical mean $\frac{1}{N_t(\gamma, i)} \sum_{s=1}^t \gamma^{t-s} X_s \mathbb{I}_{\left\{I_s=a\right\}}$  and $N^\gamma_a(t-1) =\sum_{s=1}^t \gamma^{t-s} \mathbb{I}_{\left\{A_s=a\right\}}$ is the discounted counter for arm $a$ at time $t$. The parameter $\gamma$ is tuned following Remark 3 in \cite{garivier2011upper} such as $\gamma = 1 - \mathcal{O}\left(\sqrt{\frac{K_T}{T}}\right)$.
    \item \algo{Sliding Window Risk LCB}: Inspired from the Sliding Window UCB algorithm in \cite{garivier2011upper}, the index of sliding window Risk LCB for action $a$ at time $t$ takes the following form:
    $\Index{a}{\algo{SW Risk-LCB}}{t} = \rho_{a, N^\tau_a(t-1)} - w_{a, N^\tau_a(t-1)}$
where $\rho_{a, N^\tau_a(t-1)}$ is the risk measure of the sliding window empirical mean $\frac{1}{N_t(\tau, i)} \sum_{s=t-\tau+1}^t X_s \mathbb{I}_{\left\{A_s=a\right\}}$  and $N^\tau_a(t-1) =\sum_{s=t-\tau+1}^t  \mathbb{I}_{\left\{A_s=a\right\}}$ is the sliding window counter for arm $a$ at time $t$.
    The parameter $\tau$ is tuned following Remark 9 in \cite{garivier2011upper} such as $\tau = \mathcal{O}\left(\sqrt{\frac{T \log T}{K_T}}\right)$.
    \item \algo{GLR-Risk-LCB}: Inspired from the GLR KLUCB \cite{GLRKLUCB} and TS-GLR \cite{TSGLR}, the index $\Index{a}{\algo{GLR-Risk-LCB}}{t}$ is a \algo{Risk-LCB} instance equipped with the generalized likelihood ratio for change-point detection \cite{GLR}. 
    \item The risk measure used in the experiments is the conditional value at risk defined in Example \ref{ex:cvar} with $\alpha = 0.45$.
\end{itemize}

\subsection{Synthetic Environment}
We generate a synthetic environment where the reward means originate from a Bernoulli distribution with different parameters and several local change-points.

\begin{figure}[H]
\centering
\subcaptionbox[t]{Generated piece-wise stationary Bernoulli environment with $T = 40000$, $A = 5$ and $K_T = 6$. \label{fig:Synth_env_plt}}
[0.9\linewidth]{\resizebox{0.4\textwidth}{!}{
\includegraphics{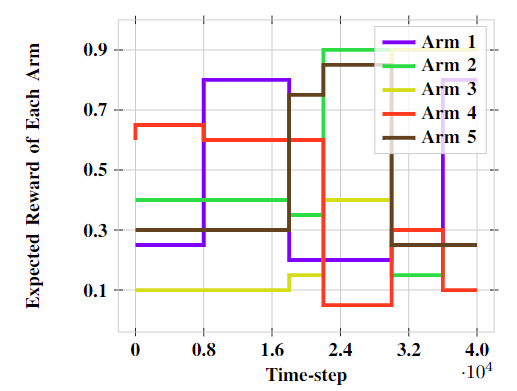}
}}%
\hfill
\\
\subcaptionbox[t]{Average cumulative regrets for different algorithms in the piece-wise stationary scenario shown in Figure \ref{fig:Synth_env_plt} over $60$ runs. \label{fig:Synth_reg_plt}}
[.9\linewidth]{\resizebox{0.4\textwidth}{!}{
\includegraphics{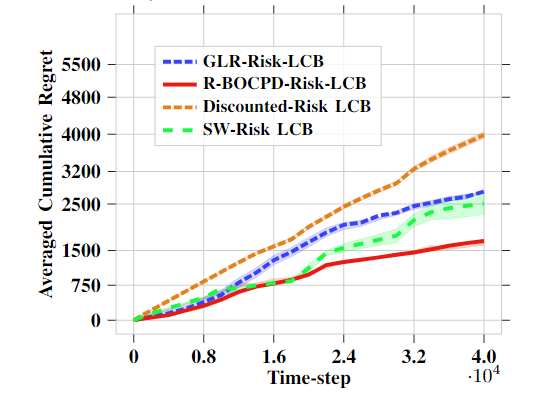}
}}%
\caption{Generated environment and cumulative regrets of risk aware MAB strategies from the synthetic dataset.}
\end{figure}    

\subsection{Real-World Dataset: Yahoo! Dataset}
We apply the previous strategies to a Yahoo! Front Page Today Module dataset \footnote{\url{https://webscope.sandbox.yahoo.com}}. This Yahoo! Dataset is a widely-used real-world dataset, collected from the Yahoo! Front Page Today Module, that serves as a valuable resource for studying piece-wise stationary risk-aware bandits. It contains a large-scale collection of anonymized user click logs, capturing diverse interactions with news' articles displayed on the front page. With a significant time span, the dataset exhibits non-stationarity as user preferences and news trends evolve over time, making it an ideal testbed for evaluating adaptive decision-making strategies.

\begin{figure}[H]
\centering
\subcaptionbox[t]{Click-through rates computed with $T = 36000$, $A = 4$ and $K_T = 10$. \label{fig:Yahoo_env_plt}}
[.9\linewidth]{\resizebox{0.4\textwidth}{!}{
\includegraphics{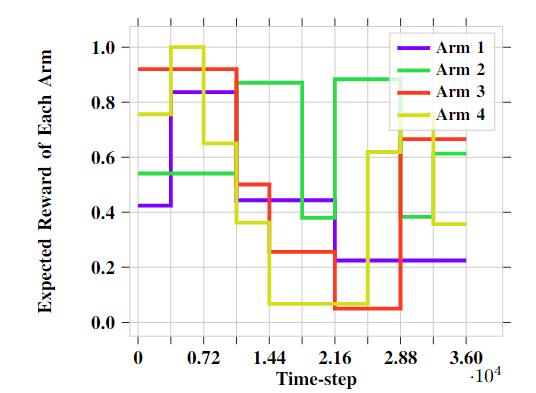}
}}%
\hfill
\subcaptionbox[t]{Averaged cumulative regrets for different algorithms in the piece-wise stationary scenario shown in Figure \ref{fig:Yahoo_env_plt} over $60$ runs. \label{fig:Yahoo_regret_plt}}
[.9\linewidth]{\resizebox{0.4\textwidth}{!}{
\includegraphics{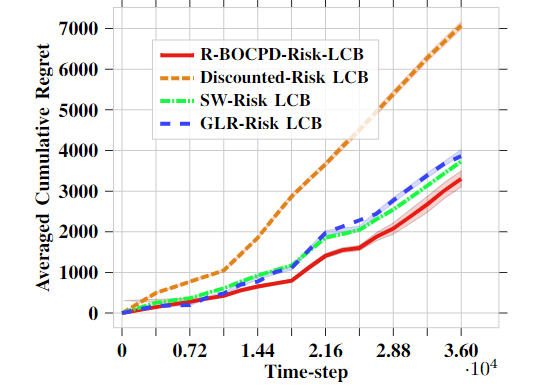}
}}%
\caption{Generated environment and cumulative regrets of risk averse MAB strategies from the Yahoo! Dataset.}
    \label{fig:Yahoo_plt}

\end{figure}

\subsection{Performance Evaluation}
We observe that the proposed algorithm $\algo{R-BOCPD-Risk-LCB}$ overcomes state-of-the-art in both synthetic and real-world environments. In the real-world setting, even while using a risk-averse measure, the huge difference in regret can be extremely drastic in practice.

\section{Limitations \& Future Work}

Solving the non-stationary risk-averse bandit problem using the Risk Lower Confidence Bound (\algo{Risk-LCB}) algorithm equipped with a Bayesian online change-point detector in a Bernoulli scenario comes with certain limitations. We express some of them as follows. 

\paragraph{Arms' Means Distribution}\algo{R-BOCPD} assumes a specific distributional form (Bernoulli in this case), which may not accurately represent the underlying reward distributions in real-world applications. If the true distributions deviate significantly from the assumed form, the change-point detection may lead to erroneous results and affect the adaptation of the LCB algorithm.

\paragraph{R-BOCPD Hyperparameters' Control}Given that \algo{R-BOCPD} is a Bayesian approach, it relies on prior assumptions and hyperparameter choices, which can influence the change-point detection and risk-averse decision-making. Selecting appropriate priors and hyperparameters can be challenging in practice, especially when there is limited prior knowledge or when the true change-points are not well-defined. 

\paragraph{Adaptive Exploration Strategy}A more elaborate choice of the \textit{Exploration-Exploitation} tradeoff parameter remains an exciting direction for future work. Balancing the constraints of an adaptive strategy(memory, complexity of online estimation) with the regret gap of a suboptimal fixed strategy would be the main challenge for algorithms of this nature.

\newpage
\bibliographystyle{plain}
\bibliography{references}

\newpage

\end{document}

%% file: math_commands.tex
\usepackage{amsmath,amsfonts,bm}

\def\eqref#1{equation~\ref{#1}}

\def\1{\bm{1}}

\DeclareMathAlphabet{\mathsfit}{\encodingdefault}{\sfdefault}{m}{sl}
\SetMathAlphabet{\mathsfit}{bold}{\encodingdefault}{\sfdefault}{bx}{n}

\DeclareMathOperator*{\argmin}{arg\,min}